\documentclass{article}
\usepackage{spconf,amsmath,graphicx}
\usepackage{url}
\usepackage{bm}
\usepackage{adjustbox}
\usepackage{color}
\usepackage{soul}
\usepackage{caption}
\usepackage{subcaption}

\title{An Anchor Learning Approach for Citation Field Learning}
\name{Zilin Yuan$^{1}$,
        Borun Chen$^{2}$, 
        Yimeng Dai$^{3}$, 
        Yinghui Li$^{1}$, 
        Hai-Tao Zheng$^{1,4,*}$, 
        Rui Zhang$^{5,*}$
        \thanks{* Corresponding author. Their E-mails are zheng.haitao@sz.tsinghua.edu.cn and rayteam@yeah.net)}}
\address{$^{1}$Shenzhen International Graduate School, Tsinghua University \\
            $^{2}$Meituan,
            $^{3}$Sapia.ai,
            $^{4}$Pengcheng Laboratory,
            $^{5}$www.ruizhang.info}
\begin{document}
%\ninept
%
\maketitle
\begin{abstract}
Citation field learning is to segment a citation string into fields of interest such as author, title, and venue. Extracting such fields from citations is crucial for citation indexing, researcher profile analysis, etc. User-generated resources like academic homepages and Curriculum Vitae, provide rich citation field information. However, extracting fields from these resources is challenging due to inconsistent citation styles, incomplete sentence syntax, and insufficient training data. To address these challenges, we propose a novel algorithm, \textbf{\textit{CIFAL}} (\underline{ci}tation \underline{f}ield learning by \underline{a}nchor \underline{l}earning), to boost the citation field learning performance. \textbf{\textit{CIFAL}} leverages the anchor learning, which is model-agnostic for any Pre-trained Language Model, to help capture citation patterns from the data of different citation styles. The experiments demonstrate that \textbf{\textit{CIFAL}} outperforms state-of-the-art methods in citation field learning, achieving a 2.68\% improvement in field-level F1-scores. Extensive analysis of the results further confirms the effectiveness of \textbf{\textit{CIFAL}} quantitatively and qualitatively.
\end{abstract}
\section{Introduction}\label{sec:intro}

Citations play a crucial role in assessing a researcher's academic accomplishments. The \textit{author} field in a citation, for instance, reveals collaborative relationships among researchers, while the \textit{title} field indicates their research interests. In this paper, our objective is to learn various citation fields, including author, title, venue, and year, from each citation. For instance, given the citation string ``\textit{Shannon, C.E., 2001. A mathematical theory of communication. ACM SIGMOBILE Mobile Computing and Communications Review, 5(1), pp.3-55.}", we aim to extract information such as the author (``\textit{Shannon, C.E.}''), title (``\textit{A mathematical theory of communication}''), venue (``\textit{ACM SIGMOBILE Mobile Computing and Communications Review}'') and year (``\textit{2001}'').  This task is commonly referred to as citation field learning, citation metadata extraction, or reference parsing in other studies~\cite{CMEviaHMM,CMEviaDNN,CMEviaFLUX}.

% \begin{figure}[h]
% \centering
% \includegraphics[width=0.5\textwidth]{pictures/intro.pdf}
% \caption{The task of citation field learning}
% \label{fig:intro}
% \end{figure}

By extracting citation fields, we can establish collaboration networks among researchers and conduct various studies, such as research community detection \cite{chen2010community}, research trends prediction \cite{behrouzi2020predicting}, and topics analysis \cite{yamamoto2006structured}. While we may obtain citation information from bibliographic databases and multiple user-generated resources, this information is vexed by several problems: the limited coverage of disciplines and researchers, inconsistent citation styles, incomplete sentence syntax, and insufficient training data. In the research community, some machine learning models~\cite{zhu2007webpage,Tang2010} and deep learning models~\cite{DBLP:conf/pakdd/ZhangDQXZ18,zhang2018pubse} have been proposed to parse the citation and have shown promising results. However, these models either rely on a substantial amount of manually labeled data or struggle with understanding unstructured information.

As it is expensive to obtain a large amount of manually labeled training data, one potential solution is to employ data enhancement techniques~\cite{I2T2I,saito-etal-2017-improving}. For instance, Zhang et al.(2018)~\cite{DBLP:conf/pakdd/ZhangDQXZ18} proposed an algorithm to augment the limited manually labeled dataset. Specifically, the algorithm automatically generates citation items based on the citation styles and meta citation data obtained from publicly available sources. Another approach is to utilize supervised citation data to fine-tune pre-trained models. By employing data enhancement techniques, we can fine-tune the pre-trained models with a larger supervised dataset, enabling them to accurately capture domain-specific and task-specific patterns. However, previous studies have indicated that using generated data in this manner may have a negative impact on transferability to downstream tasks~\cite{wang-etal-2019-tell}.

\begin{figure*}[ht]
    \centering
    \includegraphics[width=0.9\textwidth]{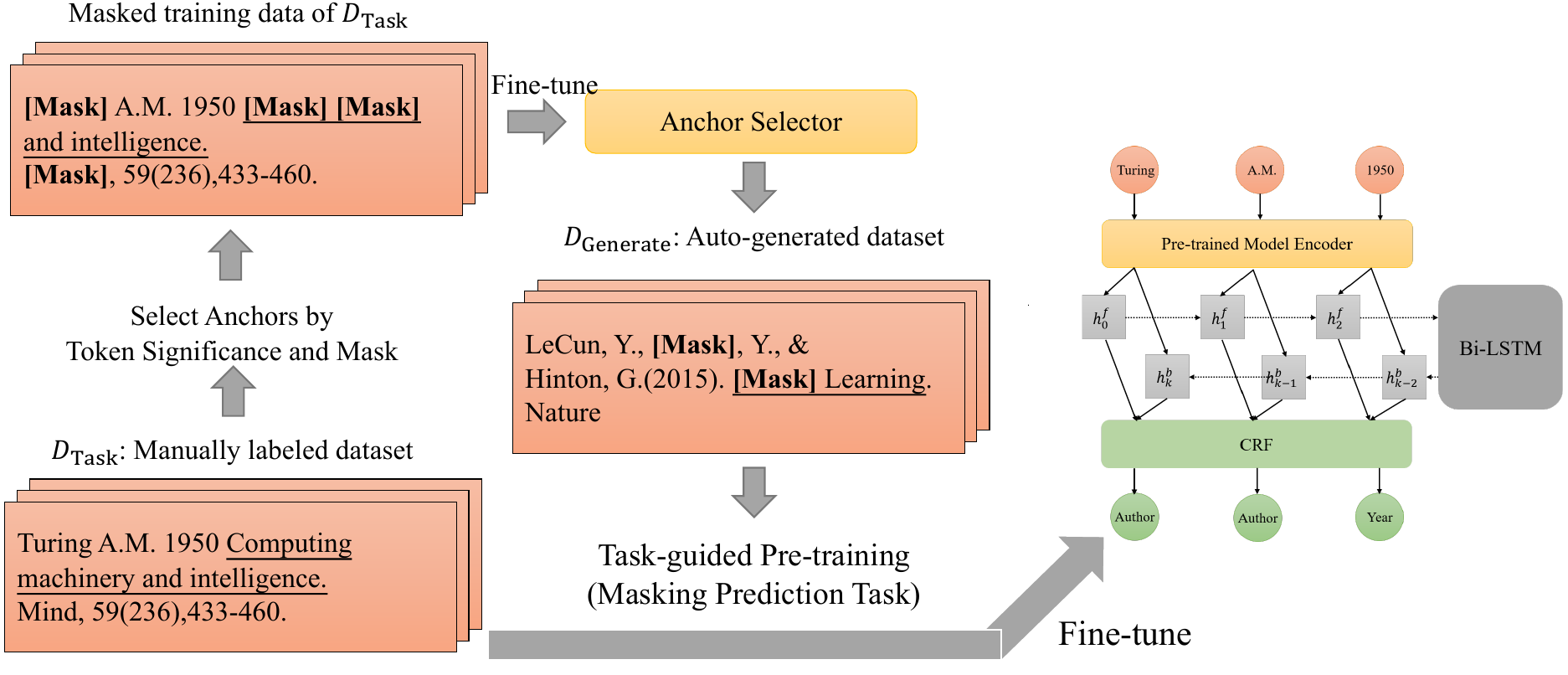}
    \caption{The anchor learning of \textbf{\textit{CIFAL}}.}
    \label{fig:CitaALA}
\end{figure*}

Therefore, we propose to leverage anchor learning to effectively capture citation patterns from the extensive generated data without compromising the transferability of pre-trained models on limited labeled data. It is model-agnostic and can be applied to any Pre-trained Language Model (PLM) by incorporating a task-guided pre-training stage between the general pre-training and fine-tuning stages~\cite{gu-etal-2020-train}. Specifically, after pre-training the PLM, the large generated data is utilized for task-guided pre-training, while the small manually labeled training dataset is used for fine-tuning. At the task-guided stage, sequence anchors, i.e., important tokens in the generated data, are masked based on our anchor learning technique. We call this masking strategy anchor masking. Then, the pre-trained model is trained to predict these anchors. The underlying idea is to facilitate the model in learning crucial citation patterns from the generated data while preserving transferability compared to direct training using labeled citations.

In summary, the main contributions of this paper are summarized as follows:

\begin{itemize}
    \item We propose a novel algorithm named \textbf{\textit{CIFAL}} (\underline{ci}tation \underline{f}ield learning by \underline{a}nchor \underline{l}earning) that leverages anchor learning to enhance the performance of citation field learning, while maintaining the transferability of pre-trained models on limited labeled data.
    \item Extensive experiments show that \textbf{\textit{CIFAL}} achieves state-of-the-art performance. Extensive model analysis and result analysis demonstrate the effectiveness of the proposed methods quantitatively and qualitatively.
\end{itemize}

The paper is organized as follows. Section \ref{method} presents our proposed methods. Section \ref{exp} reports our experimental results and Section \ref{model analysis} presents the analysis of our models. Finally, Section \ref{conclusion} concludes the paper and discusses future works.

\section{Methodology}\label{method}

\subsection{Citation Field Learning by Anchor Learning}\label{cifaka}

The pre-train-then-fine-tune paradigm has shown significant improvement for various NLP tasks~\cite{baevski-etal-2019-cloze, devlin-etal-2019-bert}. However, when faced with limited labeled data during the fine-tuning stage, models struggle to effectively capture domain-specific and task-specific patterns. To address this issue, we propose to utilize anchor learning to help capture citation patterns from a large generated dataset~\cite{DBLP:conf/pakdd/ZhangDQXZ18}. 

Figure \ref{fig:CitaALA} illustrates the anchor learning technique, which is model-agnostic and can be applied to any PLM. It involves training an anchor selector and adding a task-guided pre-training stage between the general pre-training and fine-tuning stages. We denote the large generated data as $D_{\rm Generate}$ and the small manually labeled data as $D_{\rm Task}$. Instead of randomly masking tokens in $D_{\rm Generate}$, we mask anchors in $D_{\rm Generate}$ based on an anchor selector learned on $D_{\rm Task}$.
The PLM is then task-guided pre-trained to predict these masked anchors in $D_{\rm Generate}$, allowing for the capture of citation patterns.  
Finally, the task-guided pre-trained model is fine-tuned on the citation field learning task using $D_{\rm Task}$. During the fine-tuning process, our algorithm efficiently maps tokens to the representation space and feeds these representations to the sequence labeling layer. We use Bi-LSTM-CRF to illustrate the adopted model in our method in Figure \ref{fig:CitaALA}, although other models can be used as well. Next, we'll introduce the detail of the anchor selector learned on $D_{\rm Task}$.

\subsection{Anchor Selector} 
Given an input sequence $\bm{s_g} = (\omega_1, \omega_2, \cdots, \omega_n)$ from $D_{\rm Generate}$, where $n$ is the number of tokens and $w$ represents a token, the goal of the anchor selector is to identify a set of anchors $C_g$ from $\bm{s_g}$.

Existing state-of-the-art citation field learning algorithms have shown good performance in fields such as authors and year, but they struggle to perform well in the venue field. This indicates that the language patterns in the venue are more complex and hard to capture with limited manually labeled data. As a result, we focus on training the anchor selector specifically for the venue field. In other words, the set $C_g$ will only contain tokens from the venue field. For example, in the citation item "Shannon, C.E., 2001. A mathematical theory of communication. ACM SIGMOBILE Mobile Computing and Communications Review, 5 (1), pp.3-55.", the token "ACM" can be considered as an anchor for the venue field.

We first fine-tune a PLM on $D_{\rm Task}$ to learn a basic labeling model. We use $D(\bm{v_t} \vert \bm{s_t})$ to denote the distribution of confidences for tokens from the venue field in $\bm{s_t}$, and $D(\bm{v_t}-w_i \vert \bm{s_t}-w_i)$ to denote the distribution of confidences for tokens from the venue field after removing a specific token $w_i$ from $\bm{s_t}$.
Next, we select an anchor token based on the difference between the average confidences in $D(\bm{v_t} \vert \bm{s_t})$ and the average of confidences in $D(\bm{v_t}-w_i \vert \bm{s_t}-w_i)$:
\begin{equation}\label{eq:sc}
    {\rm S}(\omega_i) = \frac{1}{k}\sum D(\bm{v_t} \vert \bm{s_t})-\frac{1}{k-1}\sum D(\bm{v_t}-w_i \vert \bm{s_t}-w_i),
\end{equation}
where $k$ is the number of tokens from the venue field.

By iteratively calculating the change of confidence after removing tokens one by one, we can determine the anchor tokens set $C_t$ for $D_{\rm Task}$. The significance of a token $\omega_i$ can be measured by its corresponding score ${\rm S}(\omega_i)$, where a higher score indicates a greater impact on the prediction and a richer source of important citation patterns information. We set the threshold $\delta = 0.05$ and the criterion for selecting anchors is:
\begin{equation}
    {\rm S}(\omega_i) > \delta,
\end{equation}
this criterion implies that when the token $\omega_i$ is removed, the average confidence of the remaining tokens significantly decreases. Therefore, tokens that meet this criterion can be considered as elements of the anchor set. 

Once the anchor sets for all sequences in $D_{\rm Task}$ are obtained, a PLM is then fine-tuned to train the anchor selector. The anchor selector is a token-level binary classifier that predicts whether a token of a sequence belongs to its anchor set or not. Subsequently, this anchor selector can be used to obtain the anchor set $C_g$ for $D_{\rm Generate}$ to perform the task-guided pre-training.

\section{Experiments}\label{exp}

\subsection{Dataset}\label{dataset}
Our experiments involve two datasets: a labeled dataset and a generated dataset. The labeled dataset is \textit{Cita}~\cite{DBLP:conf/pakdd/ZhangDQXZ18}, which consists of 3000 manually annotated citations across more than 20 disciplines. Each token in the annotated datasets is assigned a ground truth label, such as author, venue, or other. We divide the dataset into training, validation, and test sets in a ratio of 5:1:4.
Another dataset~\cite{DBLP:conf/pakdd/ZhangDQXZ18} used in this study is generated from three bibliography sources: DBLP\footnote{https://dblp.org/}, PubMed Central\footnote{https://www.ncbi.nlm.nih.gov/pmc/}, and X University\footnote{University name is hidden for anonymity} eprint library, which contains approximately 80 thousand citations. Zhang et al. (2018)~\cite{DBLP:conf/pakdd/ZhangDQXZ18} created three datasets using the aforementioned sources. They then extracted equi-size subsets from each dataset, taking into consideration the discipline distribution of \textit{Cita}. As a result, a dataset consisting of 100,026 citation items was obtained.

\subsection{Model Settings}\label{settings}

We employ the WordPiece tokenizer in BERT to tokenize each citation. For the encoder part, we select the model architecture of $\rm BERT_{BASE}$. BERT Adam is utilized during training. The initial learning rate is $5\times 10^{-5}$ and the batch size is set to 32 for both task-guided pre-training and fine-tuning. During task-guided pre-training, the model is trained for 20k steps using the generated data, which combines the anchor learning technique with random masking. In the fine-tuning stage, the model is fine-tuned for 20 epochs, and the version with the highest accuracy on the validation set is selected. To enable the models to learn from the surface features of citation items, we retain the use of uppercase letters.

\subsection{ Measurement Criteria and Models Evaluated}\label{evaluated}
In line with previous research \cite{hetzner2008simple,jagannatha2016structured}, we evaluate both field-level matching and token-level matching precision, recall, and F1-score. The primary measurement is conducted at the field-level, where each consecutive token in every field of the citation string must be correctly labeled. The token-level measurement considers each individual token, allowing partially labeled fields to receive partial credit.

We conducted an evaluation and comparison of five models, including four baseline models and our proposed model. The baseline models consist of \textbf{BioPro} \cite{chen2012bibpro}, \textbf{ParsCit}\footnote{ParsCit: \url{http://parscit.comp.nus.edu.sg/}} \cite{DBLP:conf/lrec/CouncillGK08}, \textbf{CRF} \cite{lafferty2001conditional}, and \textbf{Bi-LSTM-CRF} \cite{DBLP:journals/corr/HuangXY15}. The proposed method we tested is \textbf{\textit{CIFAL}}.

\subsection{Results}\label{results}

Table \ref{table:overall results} shows the overall results of the evaluated models\footnote{The $\dagger$ symbol in the tables denotes that the result is statistically significant.}. Based on the results of Table \ref{table:overall results}, it is evident that our methods consistently outperform all the baseline models. Specifically, \textbf{\textit{CIFAL}} achieves an improvement of up to \textbf{0.83\%} on token-level f1-score and \textbf{2.68\%} on field-level f1-score. In theory, it would be challenging to achieve significant improvement on the token-level due to the large number of tokens, where even hundreds of label changes may not have a substantial impact on performance. Conversely, the improvement on the field-level is more consistent and significant. This is reasonable as accurately predicting tokens at the field boundaries contributes significantly to the field-level performance while having minimal impact on the token-level performance.

\begin{table}[htbp]
\centering
\scriptsize
\caption{Overall Performance of Token-level and Field-level on the Cita dataset.}
\label{table:overall results}

%  \label{tab:exp_result_with_baselines}
\begin{adjustbox}{width=1.0\columnwidth,center}   
\begin{tabular}{c|ccc|ccc}

\hline
 {} & \multicolumn{3}{c|}{Token-level} & \multicolumn{3}{c}{Field-level} \\

Methods & Precision& Recall & F1 & Precision &  Recall  & F1 \\ [0.5ex]
 \hline
 % \hline
BibPro\cite{chen2012bibpro} & 93.60 & 69.87 & 80.01 & 60.76 & 50.92 & 55.41 \\

Parscit \cite{DBLP:conf/lrec/CouncillGK08} & 88.23  &  82.04  &  85.02  &  76.84  &  72.84  &  74.78 \\  

CRF \cite{lafferty2001conditional,hetzner2008simple}  & 97.14  &  96.68  &  96.91  &  85.91  &  82.73  &  84.29 \\   

Bi-LSTM-CRF \cite{DBLP:journals/corr/HuangXY15}  & 97.95 & 98.02 & 97.99 & 88.67 & 88.24 & 88.45 \\  

\hline

\textbf{\textit{CIFAL}} (Ours) & \textbf{98.77}\textsuperscript{$\dagger$} & \textbf{98.88}\textsuperscript{$\dagger$} & \textbf{98.82}\textsuperscript{$\dagger$} &\textbf{91.10}\textsuperscript{$\dagger$} &\textbf{91.16}\textsuperscript{$\dagger$} &\textbf{91.13}\textsuperscript{$\dagger$}   \\

 \hline

\end{tabular}
\end{adjustbox}

\end{table}

Table \ref{tab:each field} shows the performance of the models on each field. The table consists of four fields, each reported with f1-scores at two levels. The results indicate that \textbf{\textit{CIFAL}} achieves the highest f1-scores on the \textit{Author}, \textit{Venue}, and \textit{Title} fields. The improvement in the \textit{Venue} and \textit{Title} fields can be attributed to the fact that \textbf{\textit{CIFAL}} focuses on learning venue patterns during task-guided pre-training, and the title of most published papers have some connection to their respective venues. The improvement on \textit{Author} field indicates that the model can recognize some very challenging name cases when the obstacles to recognizing uncertain venues are clear. For instance, in book citations, the format of editors' names often closely resembles that of author names and frequently appears after the venue. Consequently, the field-level results for the \textit{Author} field exhibit significant improvement.
\begin{table}[htbp]
\centering
\scriptsize

\caption{The F1-scores for Each Field on Token-Level and Field-Level on the Cita dataset.}
\label{tab:each field}
\begin{adjustbox}{width=1.0\columnwidth,center}
\begin{tabular}{c|cc|cc|cccc}
\hline

{Methods}  &\multicolumn{2}{c}{BibPro} &\multicolumn{2}{|c|}{ParsCit} & \multicolumn{2}{c}{CRF}    \\

\hline
{Field} & Token-level  & Field-level & Token-level  & Field-level & Token-level  & Field-level\\[0.5ex]
% \hline
\hline
%               bio     parscit             crf               bi-lstm-crf
\textit{Author} & 80.41  & 22.61 & 87.61  &  81.24 & 99.04  &  83.23   \\
\textit{Year}   & 82.88 & 82.59& 93.76  &  97.71 & \textbf{99.46}\textsuperscript{$\dagger$}  &  \textbf{99.40}\textsuperscript{$\dagger$}    \\
\textit{Venue}  & 72.03 & 59.99 & 62.84  &  55.59 & 86.64  &  76.42      \\
\textit{Title}  & 81.70 & 56.14 & 88.45  &  62.50 & 95.64  &  75.95     \\
\hline
Average         & 80.01 & 55.41 & 85.02  &  74.78 & 96.91  &  84.29      \\
\hline
\end{tabular}
\end{adjustbox}
 \label{tab:exp_result_detailed}
\end{table}
\begin{table}[h]
\centering
\scriptsize

\begin{tabular}{c|cc|cccc}
\hline

{Methods} & \multicolumn{2}{c}{Bi-LSTM-CRF}  &\multicolumn{2}{|c}{\textbf{\textit{CIFAL}} (Ours)}  \\
\hline
{Field} & Token-level  & Field-level & Token-level  & Field-level \\[0.5ex]
% \hline
\hline
\textit{Author} & 99.46 & 89.17  & \textbf{99.74}\textsuperscript{$\dagger$} & \textbf{91.29}\textsuperscript{$\dagger$} \\
\textit{Year}   & 96.90 & 96.07 & 99.29 & 99.08 \\
\textit{Venue}  & 91.08 & 84.96 & \textbf{94.32}\textsuperscript{$\dagger$} & \textbf{87.26}\textsuperscript{$\dagger$}  \\
\textit{Title}  & 97.33 & 83.77 & \textbf{98.54}\textsuperscript{$\dagger$} & \textbf{86.61}\textsuperscript{$\dagger$} \\
\hline
Average         & 97.99 & 88.45 & \textbf{98.82}\textsuperscript{$\dagger$}& \textbf{91.13}\textsuperscript{$\dagger$} \\
\hline
\end{tabular}
% \end{adjustbox}
 \label{ours}
\end{table}

\section{Model Analysis}\label{model analysis}

\subsection{Ablation Study}
The task-guided pre-training and the anchor masking technique may both contribute to our state-of-the-art performances on the citation field learning task. Hence, we make modifications to the \textbf{\textit{CIFAL}} model by employing different settings for task-guided pre-training and anchor masking. We test 4 versions of modified models: 1) \textbf{Fine-tune}: The \textbf{\textit{CIFAL}} model without task-guided pre-training and only be fine-tuned on the manually labeled data; 2) \textbf{Anchor Masking}: The \textbf{\textit{CIFAL}} model with task-guided pre-training by anchor masking; 3) \textbf{Attention Masking}: The \textbf{\textit{CIFAL}} model with task-guided pre-training by attention masking, and it should be noted that the attention masking strategy we utilize here is based on the attention weights of BERT Encoder layers.; 4) \textbf{Random Masking}: The \textbf{\textit{CIFAL}} model with task-guided pre-training by random masking.

We present the f1-scores on field-level of each field in Table \ref{tab:ablation_results}. The results of the ablation experiments demonstrate that the impact of task-guided pre-training with anchor learning technique is greater than any other method. Specifically, the venue field achieves an f1-score of 87.26\%, which significantly outperforms the other modified models. This outcome highlights the effectiveness of the anchor learning technique in improving the performance of the venue field.

\begin{table}[htbp]
\centering
\scriptsize
\caption{Each Field and Overall F1-scores on Field-level of modified models.}
\begin{tabular}{c|c|c|c|c|c}
\hline
{Methods} & \textit{Author} & \textit{Title} & \textit{Venue} & \textit{Year} & Overall\\[0.5ex]
\hline
Fine-tune &91.06 	&86.42 	&86.04 	&\textbf{99.39} 	&90.81       \\
% \hline
% \hline
Anchor Masking &91.29 	&86.61 	&\textbf{87.26} 	&99.08 	&\textbf{91.13}  \\
% \hline
Attention Masking & \textbf{91.51} & 86.94 & 86.60 & 99.27 & 91.04  \\
% \hline
Random Masking &91.24 &\textbf{87.07} &86.22 &99.13 &90.99  \\
\hline
\end{tabular}
 \label{tab:ablation_results}
\end{table}

\subsection{Effect of Anchor Learning}

To prove that our anchor learning technique does select the anchor set of the input sequence for the target field, we present some examples and count the tokens that appear most frequently in the anchor set obtained by applying our strategy on the generated data.

Considering the citation item ``\textit{Voelcker, J ( 2013), Communications and Navigation, IEEE Spectrum, pages 2574-2583.}", when we use the anchor learning technique for this input, the model will mask the tokens ``\textit{IEEE}" and ``\textit{Spectrum}" for task-guided pre-training, where the two masked tokens just belong to the venue field. It proves that anchor learning does select the important tokens for the venue field. 
Additionally, we conducted a statistical analysis on the generated data and got the twelve most frequently masked tokens. Among these tokens, ``\textit{Journal}" is the most frequently masked token and it occurs 57,195 times in total. This result aligns with our expectations, as ``\textit{Journal}" is a commonly used term in most venues. Other high-frequency tokens such as "Conference" and "Research" are also prevalent in the venue field.

Based on the example and the statistical analysis, we can see the effectiveness of our anchor learning technique. In this way, the task-guided pre-training can better capture the task-specific language pattern.

\section{Conclusion}\label{conclusion}
In this paper, we studied the problem of citation field learning and proposed a novel method called \textbf{\textit{CIFAL}}. Our approach utilizes anchor learning techniques to effectively improve the performance in this domain, while also ensuring the transferability of pre-trained models. The experimental results demonstrate that our model significantly outperforms existing state-of-the-art solutions. Additionally, we conducted further ablation studies to evaluate the impact of anchor learning in our proposed methods. To further enhance our approach, future work could focus on improving the semantic representation capacity within the citation field in \textbf{\textit{CIFAL}}.

\vfill\pagebreak
\clearpage

\small
\bibliographystyle{IEEEbib}
\bibliography{strings,refs}

\end{document}